\documentclass[12pt,onecolumn,conference, draftclsnofoot]{IEEEtran}
\IEEEoverridecommandlockouts
\usepackage{cite}
\usepackage{amsmath,amssymb,amsfonts}
\usepackage{algorithmic}
\usepackage{graphicx}
\usepackage{textcomp}
\usepackage{xcolor}
\newsavebox{\largestimage}

\usepackage{graphicx}
\usepackage{comment}
\usepackage{amsmath,amssymb}
\usepackage{color}
\usepackage{relsize}
\usepackage{textcomp}

\usepackage{makecell}
\usepackage{tabularx}
\usepackage{array}
\usepackage{tabularx}
\usepackage{multirow}
\usepackage{float}
\usepackage{pifont}%
\usepackage{listings}
\usepackage{color}
\usepackage{mdframed} 
\usepackage{arydshln}
\usepackage{mathtools}
\usepackage{subcaption}
\usepackage{titlesec}
\usepackage[normalem]{ulem}
\usepackage{url}
\usepackage{hyperref}

\usepackage[normalem]{ulem}
\useunder{\uline}{\ulined}{}%
\DeclareUrlCommand{\bulurl}{}

\DeclareUrlCommand\ULurl{%
  \renewcommand\UrlLeft{\uline\bgroup}%
  \renewcommand\UrlRight{\egroup}}

\titlespacing{\paragraph}{%
  0pt}{%
  0.2\baselineskip}{%
  0.4em}%

\definecolor{light-gray}{gray}{0.95} %

\newcommand{\cmark}{\ding{51}}%
\newcommand{\xmark}{\ding{55}}%
\def\BibTeX{{\rm B\kern-.05em{\sc i\kern-.025em b}\kern-.08em
    T\kern-.1667em\lower.7ex\hbox{E}\kern-.125emX}}

\begin{document}

\title{RADIATE: A Radar Dataset for Automotive Perception in Bad Weather\\
\thanks{}
}

\author{
\IEEEauthorblockN{
Marcel Sheeny\IEEEauthorrefmark{1},
Emanuele De Pellegrin\IEEEauthorrefmark{1},
Saptarshi Mukherjee\IEEEauthorrefmark{1},\\
Alireza Ahrabian\IEEEauthorrefmark{1},
Sen Wang\IEEEauthorrefmark{1},
Andrew Wallace\IEEEauthorrefmark{1}
}
\IEEEauthorblockA{\IEEEauthorrefmark{1}Institute of Sensors, Signals and Systems, Heriot-Watt University}
}

\maketitle

\begin{abstract}
Datasets for autonomous cars are essential for the development and benchmarking of perception systems. However, most existing datasets are captured with camera and LiDAR sensors in good weather conditions. In this paper, we present the RAdar Dataset In Adverse weaThEr (RADIATE), aiming to facilitate research on object detection, tracking and scene understanding using radar sensing for safe autonomous driving. RADIATE includes 3 hours of annotated radar images with more than 200K labelled road actors in total, on average about 4.6 instances per radar image. It covers 8 different categories of actors in a variety of weather conditions (e.g., sun, night, rain, fog and snow) and driving scenarios (e.g., parked, urban, motorway and suburban), representing different levels of challenge. To the best of our knowledge, this is the first public radar dataset which provides high-resolution radar images on public roads with a large amount of road actors labelled. The data collected in adverse weather, e.g., fog and snowfall, is unique. Some baseline results of radar based object detection and recognition are given to show that the use of radar data is promising for automotive applications in bad weather, where vision and LiDAR can fail. RADIATE also has stereo images, 32-channel LiDAR and GPS data, directed at other applications such as sensor fusion, localisation and mapping. The public dataset can be accessed at
\url{http://pro.hw.ac.uk/radiate/}.

\end{abstract}

\begin{figure*}[t]
\begin{center}
\includegraphics[width=0.9\textwidth]{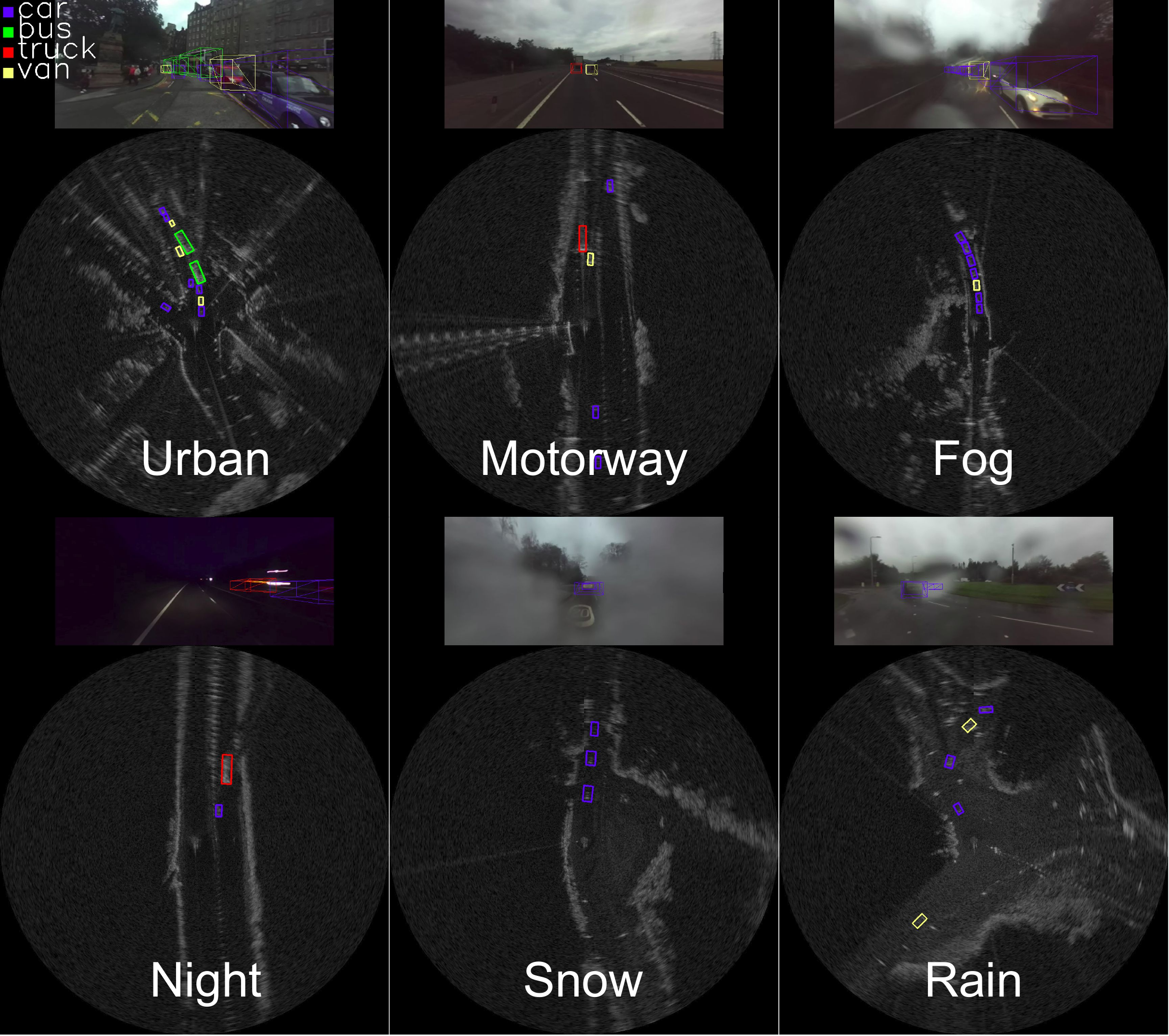}
\end{center}
\caption{Examples from \textbf{RADIATE}. This dataset contains radar, stereo camera, LiDAR and GPS data. It was collected in various weather conditions and driving scenarios with 8 categories of annotated objects.}\label{fig:radiate_data}
\end{figure*}

\section{Introduction}\label{intro}

 Autonomous driving research and development rely heavily on the use of public datasets in the computer vision and robotics communities \cite{Geiger2013IJRR, nuscenes2019, sun2019scalability}. Camera and LiDAR are the two primary perceptual sensors that are usually adopted. However, since these are visible spectrum sensors, the data is affected dramatically by bad weather, causing attenuation, multiple scattering and turbulence \cite{pfennigbauer2014online, SDV18, bijelic2018benchmark, kutila2018automotive,WalHalBul2020}.
 On the other hand, a radar sensor is known to be more robust in adverse conditions \cite{daniel2017low, norouzian2019experimental, norouzian2019rain}. However, there are few public radar datasets for automotive applications, especially in bad weather and with object annotation.

In this paper, we present the RAdar Dataset In Adverse weaThEr (RADIATE) for perceptual tasks in autonomous driving and driver assistance. It includes a mixture of weather conditions and driving scenarios, representing different levels of challenge. A high-resolution 79 GHz 360{\textdegree} radar is chosen as the main sensor for object annotation. RADIATE includes 5 hours of radar images in total, 3 hours of which are fully annotated. This gives RADIATE more than 200K labelled object instances with 8 categories of road actors (i.e., car, van, bus, truck, motorbike, bicycle, pedestrian and a group of pedestrians). RADIATE also has stereo camera, LiDAR and GPS data collected simultaneously. Some examples from RADIATE are shown in Figure \ref{fig:radiate_data}.
\vspace{2em}

Our work makes the following main contributions:
\begin{itemize}
    \setlength\itemsep{-0.4em}
    \item To the best of our knowledge, RADIATE is the first public radar dataset which includes a large number of labelled road actors on public roads.
    \item It includes multi-modal sensor data collected in challenging weather conditions, such as dense fog and heavy snowfall. Camera, LiDAR and GPS data are also provided for all sequences.
    \item As an example of a use case, we demonstrate that RADIATE can be used for robust vehicle detection in adverse weather, when optical sensors (camera and LiDAR) fail.
\end{itemize}

\section{Related Work}

There are many publicly available datasets for research into perception for autonomous and assisted driving. The most popular is the KITTI dataset \cite{Geiger2013IJRR}, using cameras and a Velodyne HDL-64e LiDAR to provide data for several tasks, such as object detection and tracking, odometry and semantic segmentation. Although widely used as a benchmark, data was captured only in good weather and appears now as rather small scale.
Waymo \cite{sun2019scalability} and Argo \cite{Chang_2019_CVPR} are automotive datasets which are larger than KITTI and provide more data variability. Some data was collected in the rain and at night although adverse weather is not their main research focus. Foggy Cityscapes \cite{SDV18} developed a synthetic foggy dataset aimed at object recognition tasks but only images are provided.
All these datasets use only optical sensors.

\begin{table}[t]
 \caption{Comparison of RADIATE with public automotive datasets that use radar sensing.}
    \begin{center}
    \resizebox{\columnwidth}{!}{
    \begin{tabular}{l|c|ccc|cccc|ccc|c}
    \hline
    Dataset          & Size  & Radar                      & Lidar & Camera & Night &Fog & Rain & Snow & \makecell{Object\\ Detection} & \makecell{Object\\ Tracking}  & Odometry                               & \makecell{3D\\ Annotations} \\
    \hline \hline
    nuScenes \cite{nuscenes2019}        & Large & \makecell{\cmark \\(Sparse Point Cloud)}   & \cmark   & \cmark    & \cmark  & \xmark& \cmark  & \xmark   & \cmark              & \cmark                             & \xmark                                     & \cmark            \\ \hline
    Oxford Radar RobotCar \cite{RadarRobotCarDatasetICRA2020}   & Large & \makecell{\cmark \\ (High-Res Radar Image)} & \cmark   & \cmark    &\cmark &\xmark  & \cmark   & \xmark   & \xmark               & \xmark                                & \cmark                                    & \xmark            \\ \hline
    MulRan \cite{gskim-2020-mulran}           & Large & \makecell{ \cmark  \\(High-Res Radar Image)}       & \cmark    & \xmark & \xmark&\xmark & \xmark   & \xmark   & \xmark               & \xmark                            & \cmark                                    & \xmark          \\ \hline
    Astyx \cite{meyer2019automotive}            & Small & \makecell{\cmark\\ (Sparse Radar Point Cloud)} & \cmark   & \cmark & \xmark   & \xmark  & \xmark   & \xmark   & \cmark              & \xmark                                & \xmark                                     & \cmark           \\ \hline
    RADIATE (Ours)          & Large &\makecell{ \cmark\\ (High-Res Radar Image)} & \cmark   & \cmark    & \cmark & \cmark & \cmark  & \cmark  & \cmark              & \cmark                             & \cmark
    & \makecell{\xmark\\ (Pseudo-3D)} \\
    \hline
    \end{tabular}\label{tab:comparison}}
    \end{center}
   
    \end{table}

Radar, on the other hand, provides a sensing solution that is more resilient to fog, rain and snow. It usually provides low-resolution images, which makes it very challenging for object recognition or semantic segmentation. Current automotive radar technology relies, usually on the Multiple Input Multiple Output (MIMO) technique, which uses several transmitters and receivers to measure the direction of arrival (DOA) \cite{tiradar}. Although this is inexpensive, current configurations lack azimuth resolution, e.g. the cross-range image of a commercial radar with 15{\textdegree} angular resolution is around 10 meters at 20-meter distance. This means that a radar image provides insufficient detail for object recognition or deatiled scene mapping. Scanning radar measures at each azimuth using a moving antenna, providing better azimuth resolution. This type of sensor has recently been developed to tackle radar based perception tasks for automotive applications \cite{daniel2017low, SheWalWan2020, sheeny2020radio}.

For most datasets which provide radar data for automotive applications, the radar is used only as a simple detector, e.g. NuScenes \cite{nuscenes2019}, to give sparse 2D point clouds. Recently, the Oxford Robotcar \cite{RadarRobotCarDatasetICRA2020} and MulRan datasets \cite{gskim-2020-mulran} provided data collected from a scanning Navtech radar in various weather conditions. However, they do not provide object annotations as the data was designed primarily for Simultaneous Localisation and Mapping (SLAM) and place recognition in long-term autonomy. The Astyx dataset \cite{meyer2019automotive} provides denser data (compared to current MIMO technology) but with only about $500$ frames annotated. It is also limited in terms of weather variability. Table \ref{tab:comparison} compares existing public automotive radar datasets with RADIATE.
To summarise, although research into radar perception for autonomous driving has recently been gaining popularity \cite{Major_2019_ICCV, Sless_2019_ICCV,wang2020rodnet}, there is no radar dataset with a large set of actor annotations publicly available. We hope the introduction of RADIATE can boost autonomous driving research in the community.

\section{The RADIATE Dataset}

\begin{figure*}[t]
\begin{center}
        \includegraphics[width=1.0\linewidth]{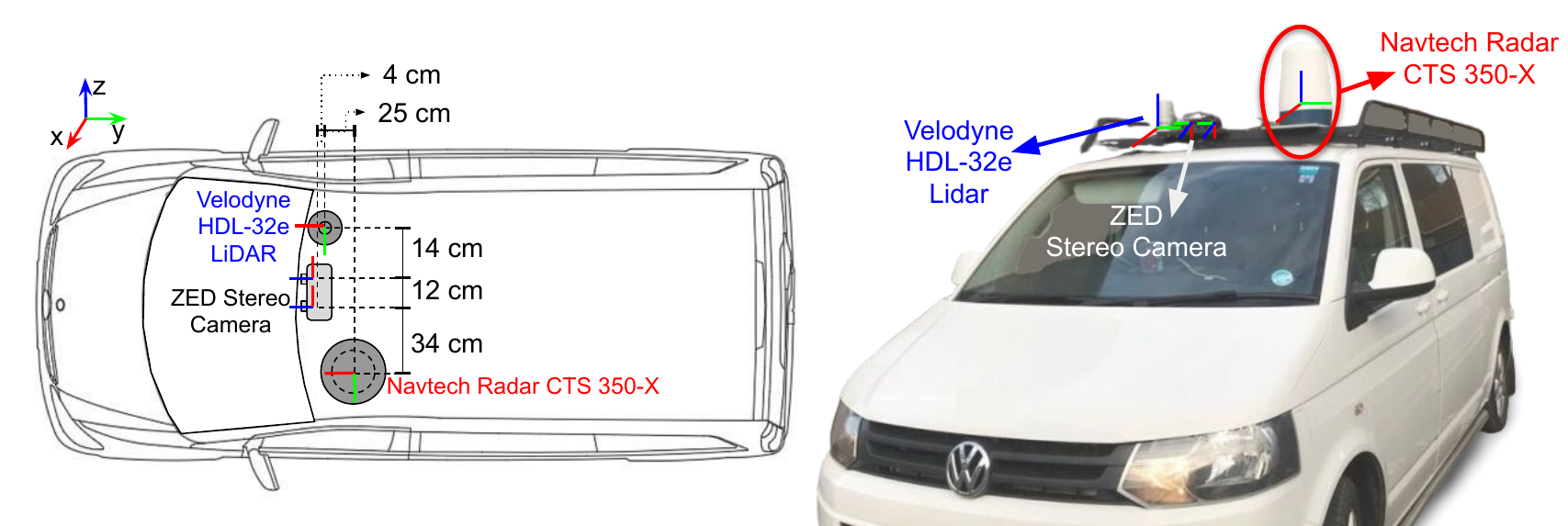}
    \end{center}

    \caption{Sensor setup for data collection.}\label{fig:sensors_setup}
\end{figure*}

\begin{figure}[h]
    \begin{center}
    \includegraphics[width=0.7\textwidth]{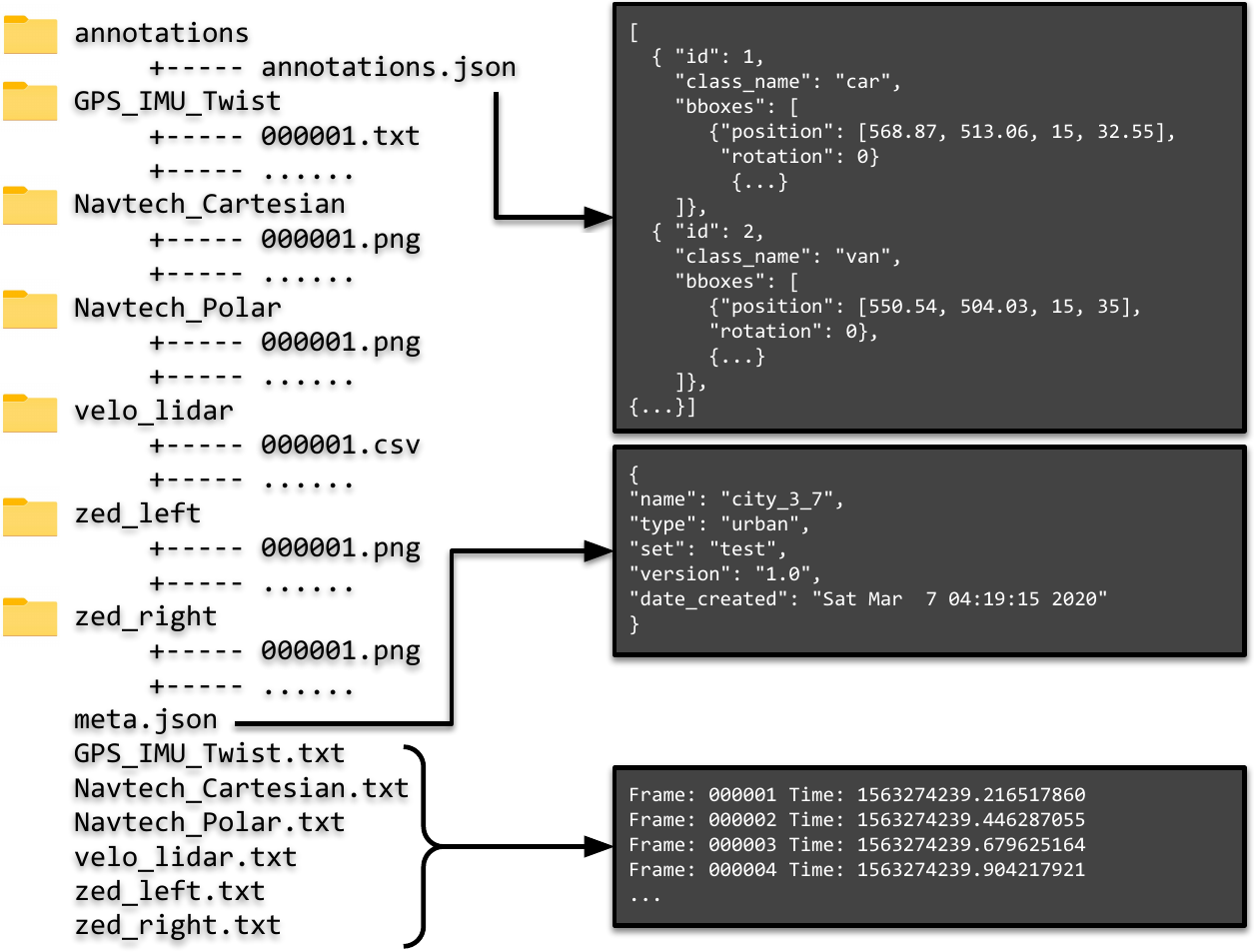}
    \end{center}
    \caption{Folder tree, annotation, metadata and timestamp structure of each sequence}
    \label{fig:folder_tree}
\end{figure}

The RADIATE dataset was collected between February 2019 and February 2020. The data collection system was created using the Robot Operating System (ROS) \cite{ros}. From the ''rosbag'' created by ROS, we extracted sensor information, each with its respective timestamp. Figure \ref{fig:folder_tree} shows the folder structure used for the dataset.
To facilitate access, a RADIATE software development kit (SDK) is released for data calibration, visualisation, and pre-processing.

\subsection{Perception Sensors}

RADIATE includes radar, LiDAR and stereo cameras. Figure \ref{fig:sensors_setup} shows the sensor setup and the extrinsic configuration on our vehicle. To support sensor fusion, the sensors are calibrated (details in Section \ref{sec:sensor_calibration}).

\paragraph{Stereo Camera} An off-the-shelf ZED stereo camera is used. It is set at $672 \times 376$ image resolution at 15 frames per second for each camera. It is protected by a waterproof housing for extreme weather. The images can be seriously blurred, hazy or fully blocked due to rain drops, dense fog or heavy snow, respectively. Some examples are shown in Figure \ref{fig:radiate_data}.

\paragraph{LiDAR} A 32 channel, 10Hz, Velodyne HDL-32e LiDAR \cite{velodyne} is used to give 360{\textdegree} coverage. Since the LiDAR signal can be severely attenuated and reflected by intervening fog or snow \cite{WalHalBul2020}, the data can be missing, noisy and incorrect.

\paragraph{Radar} RADIATE adopts the Navtech CTS350-X \cite{navtech_radar} radar. It is a scanning radar which provides 360{\textdegree} high-resolution range-azimuth images. It has 100 meters maximum operating range with 0.175m range resolution, 1.8{\textdegree} azimuth resolution and 1.8{\textdegree} elevation resolution, Currently, it does not provide Doppler information.

\subsection{Sensor Calibration} \label{sec:sensor_calibration}

Sensor calibration is required for multi-sensor fusion, feature and actor correspondence. The intrinsic parameters and distortion coefficients of the stereo camera are calibrated using the Matlab camera calibration toolbox \cite{matlabtool}. Then, rectified images can be generated to calculate depths. In terms of extrinsic calibration, the radar sensor is chosen as the origin of the local coordinate frame as it is the main sensor. The extrinsic parameters for the radar, camera and LiDAR are represented as 6 degree-of-freedom transformations (translation and rotation). They are performed by first explicitly measuring the distance between the sensors, and then fine-tuned by aligning measurements between each pair of sensors. The sensor calibration parameters are provided in a {\tt yaml} file. The sensors operate at different frame rates and  we simply adopt the time of arrival of the data at each sensor as the timestamp.

\subsection{Data Collection Scenarios}

We collected data in 7 different scenarios, i.e., sunny (parked), sunny/overcast (urban), overcast (motorway), night (motorway), rain (suburban), fog (suburban) and snow (suburban).

\paragraph{Sunny (Parked)}
In this scenario the vehicle was parked at the side of the road, sensing the surrounding actors passing by. This is intended as the easiest scenario for object detection, target tracking and trajectory prediction. This was collected in sunny weather.

\paragraph{Sunny/Overcast (Urban)}
The urban scenario was collected in the city centre with busy traffic and dense buildings. This is challenging since many road actors appear. The radar collected in this scenario is also cluttered by numerous reflections from non-road actors, such as trees, fences, bins and buildings, and multi-path effects, increasing the challenge. Those sequences were collected in sunny and overcast weather.

\paragraph{Overcast (Motorway)}
The motorway scenario was captured on the Edinburgh city bypass. This can be considered as relatively easy since most of the surrounding dynamic actors are vehicles and the background is mostly very similar. This scenario was collected in overcast weather.

\paragraph{Night (Motorway)}
We collected data during night in a motorway scenario. Night is an adverse scenario for passive optical sensors due to lack of illumination. Lidar and radar were expected to behave well since they are active sensors and do not depend on an external light source.

\paragraph{Rain (Suburban)}
We collected 18 minutes of data in substantial rain. The collection took place in a suburban scenario close to the university campus.

\paragraph{Fog (Suburban)}
We found foggy data challenging to collect. Fog does not happen very often in most places and it is very hard to predict when it will occur and how dense it will be. In practice, we collected foggy data opportunistically when parked and driving in suburban areas.

\begin{figure}
    
    \begin{minipage}{0.7\textwidth}
        \begin{center}
        \includegraphics[width=1.0\textwidth]{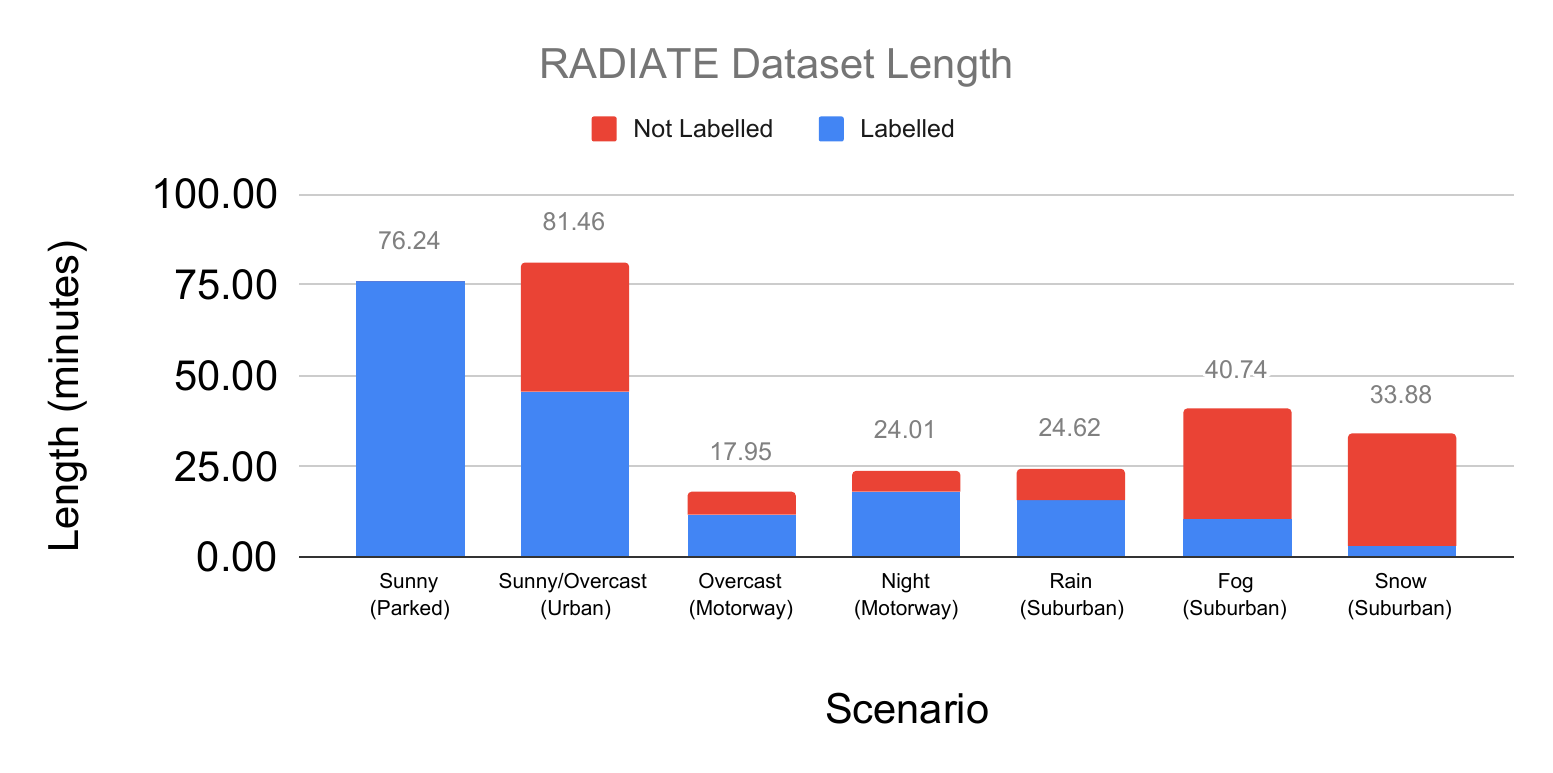} %
    \end{center}
    \caption{Dataset length for driving scenarios and weather conditions (in minutes).}\label{fig:dataset_length}
    \end{minipage}\hfill
    \begin{minipage}{0.285\textwidth}
        \begin{center}
        \includegraphics[width=1.0\textwidth]{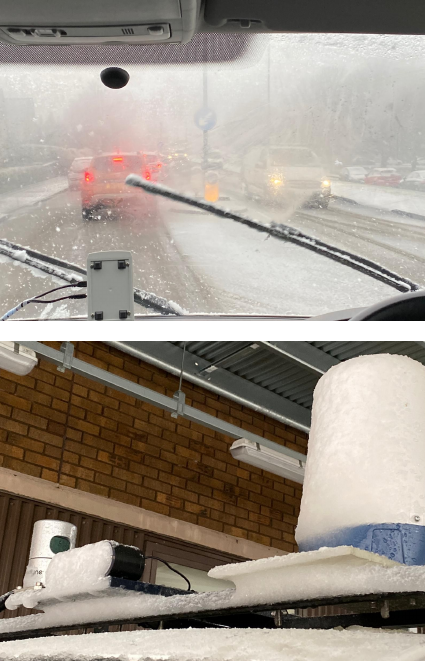} %
    \end{center}
    \begin{center}
    \caption{Sensors covered in snow}\label{fig:sensors_snow}
    \end{center}
    \end{minipage}
\end{figure}

\paragraph{Snow (Suburban)}
RADIATE includes 34 minutes of data in snow, in which 3 minutes are labelled. Snowflakes are expected to interfere with LiDAR and camera data, and can also affect radar images to a lesser extent. Moreover, heavy snowfall can block the sensors within 3 minutes of data collection (see Figure \ref{fig:sensors_snow}). The data in snow was collected in a suburban scenario.

\begin{figure*}[t!]
    \begin{center}
        \includegraphics[width=\textwidth]{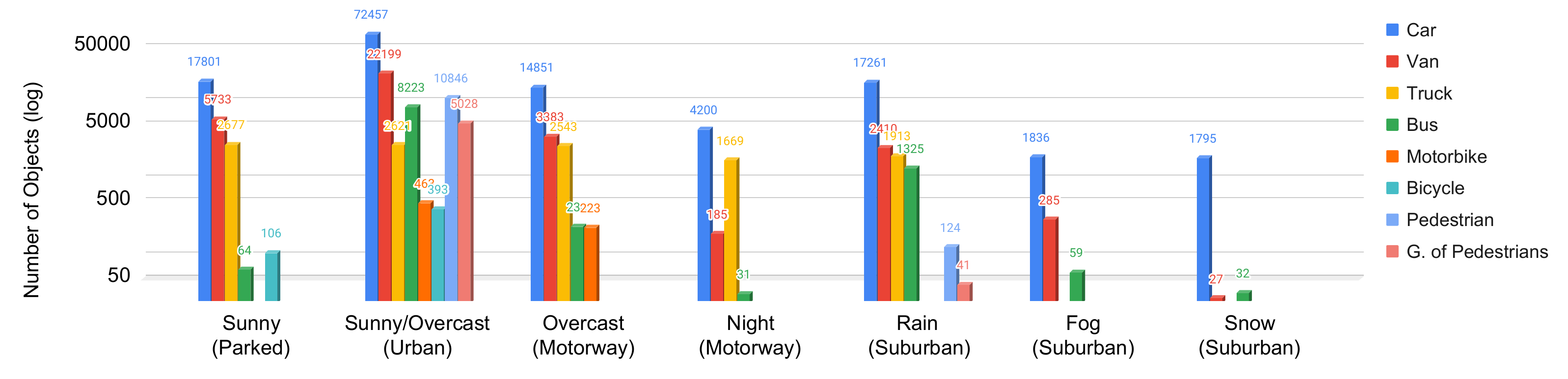}
    \end{center}
    \caption{Category distribution for each scenario.}\label{fig:class_distribution}
\end{figure*}

\subsection{Labelling}\label{sec:labelling}

Labelling radar images is challenging because it is not easy for a human to recognise objects in radar data. This means most existing annotation tools which are designed for a single sensor \cite{cvat, vatic, labelimg} are inadequate. Therefore, a new annotation tool was developed to automatically correlate and visualise multiple sensors through sensor calibration.

Eight different road actors are labelled in RADIATE: \textit{cars, vans, trucks, buses, motorbikes, bicycle, pedestrian and group of pedestrians}. 2D bounding boxes were annotated on radar images after verifying the corresponding camera images. Each bounding box is represented as (x,y,width,height,angle), where (x,y) is the upper-left pixel locations of the bounding box, of given width and height and counter-clockwise rotation angle. To achieve efficient annotation, the CamShift tracker was used \cite{allen2004object}.  

In total, RADIATE has more than $200$K bounding boxes over $44$K radar images, an average 4.6 bounding boxes per image. The class distribution for each driving scenario is illustrated in Figure \ref{fig:class_distribution}. In Figure \ref{fig:dataset_length}, the length and the total number of objects in each scenario are given. 

\begin{figure*}[t!]
        \begin{center}
        \includegraphics[width=\textwidth]{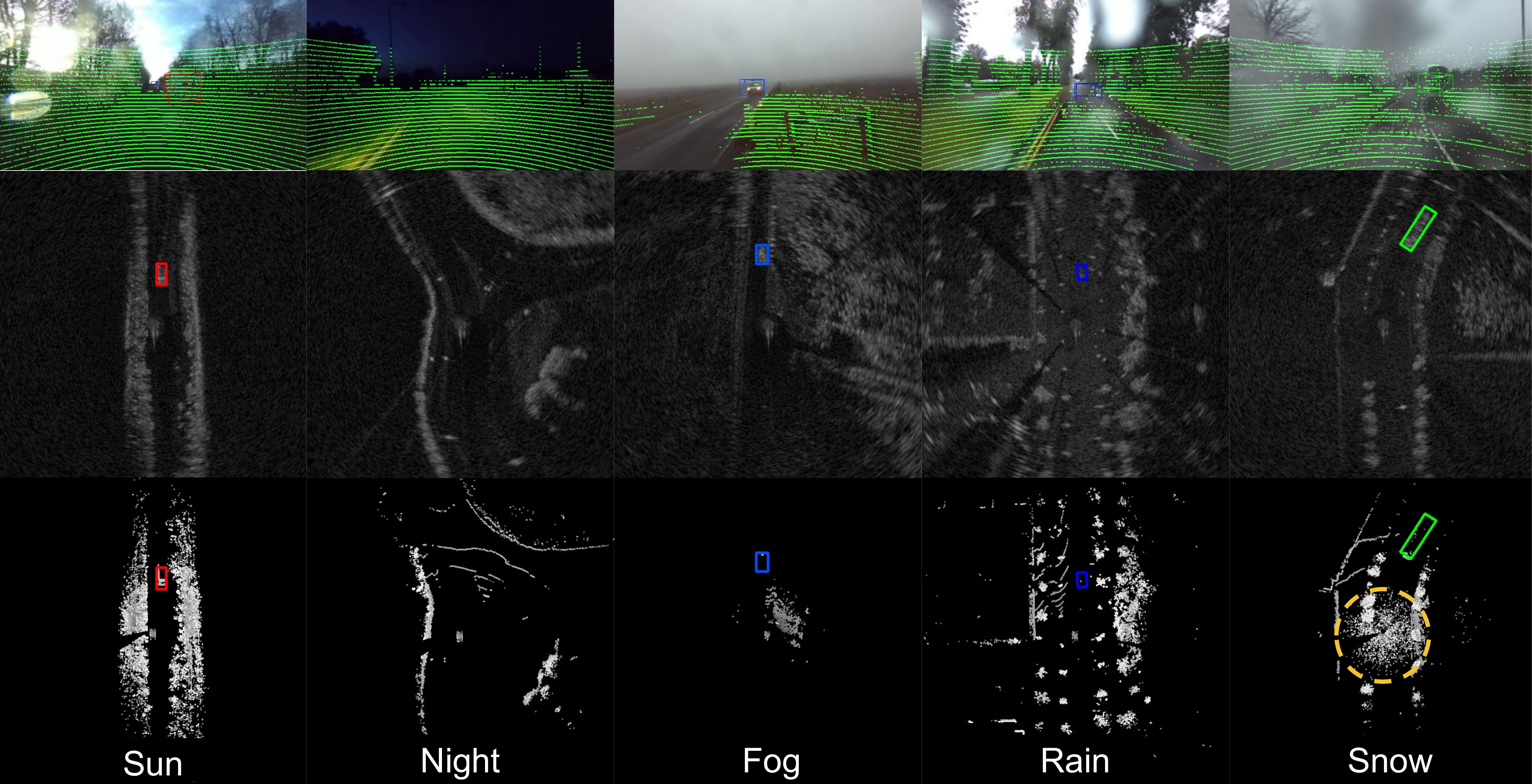}
        \end{center}
        \caption{Data in various weather conditions. \textbf{Top:} Image with LiDAR points projected. \textbf{Middle:} Radar with objects annotated. \textbf{Bottom}: LiDAR with objects projected from radar annotation. Note both image and LiDAR images are degraded in fog, rain and snow. The yellow circles encloses false LiDAR points caused by snow flakes.}\label{fig:lidar_fog}
\end{figure*}

\section{Example of Use: Radar based Vehicle Detection}

One of the key perception capabilities of autonomous vehicles is to detect and recognise road actors for safe navigation and decision making. To the best of our knowledge, there is no existing work on radar based object detection for autonomous driving in extreme weather. Therefore, we present as a use case the first baseline results using RADIATE.
Figure \ref{fig:lidar_fog} shows some examples of our data collected in adverse weather, in which trained networks fail to recognise actors in the optical images. As shown in Figure \ref{fig:lidar_fog}. The radar images are less affected so we would hypothesise that robust object detection is more likely in all conditions.

Drawn from the data acquired in several weather conditions, RADIATE includes 3 exemplary data sets that can be used for evaluation for vehicle detection:
\begin{itemize}
    \item \textbf{Training set in good weather}: This only contains data in good weather, sunny or overcast. It was created to validate whether sensors are capable to adapt from good weather to bad weather conditions.
    \item \textbf{Training set in good and bad weather}: This additional set includes data from both good and bad weather conditions (night, rain, fog and snow). This was used to develop algorithms for all weathers.
    \item \textbf{Test set}: This test set includes data from both good and bad weather conditions and is used for evaluation and benchmarking.
\end{itemize}

As a first baseline, we have performed evaluation of vehicle detection from single images. We defined a vehicle as one of the following classes: car, van, truck, bus, motorbike and bicycle. In Table \ref{tab:set_size} are shown the number of radar images and vehicles used to train and test the network.

\begin{table}[htbp]
\caption{Number of images for each set defined.}
\begin{center}
\begin{tabular}{l|c|c}
\hline
{} & \#Radar Images&  \#Vehicles   \\
\hline
\hline
Training Set in Good Weather      &     23,091 & 106,931  \\
\hline
Training Set in Good and Bad Weather &      9,760 & 39,647   \\
\hline
Test Set                                              &     11,289 & 42,707  \\
\hline
\hline
Total                                              &     44,140 & 147,005 \\
\hline
\end{tabular}
\label{tab:set_size}
\end{center}
\end{table}

\subsection{Radar based Vehicle Detection in the Wild}

We adopted the popular Faster R-CNN \cite{ren2015faster} architecture to demonstrate the use of RADIATE for radar based object detection. Two modifications were made to the original architecture to better suit radar detection:

\begin{itemize}
    \item Pre-defined sizes were used for anchor generation because vehicle volumes are typically well-known and radar images provide metric scales, different from camera images. 
    \item We modified the Region Proposal Network (RPN) from Faster R-CNN to output the bounding box and a rotation angle which the bounding boxes are represented by {\it x, y, width, height, angle}.
\end{itemize}

\begin{figure}
        \begin{center}
        \includegraphics[width=0.5\textwidth]{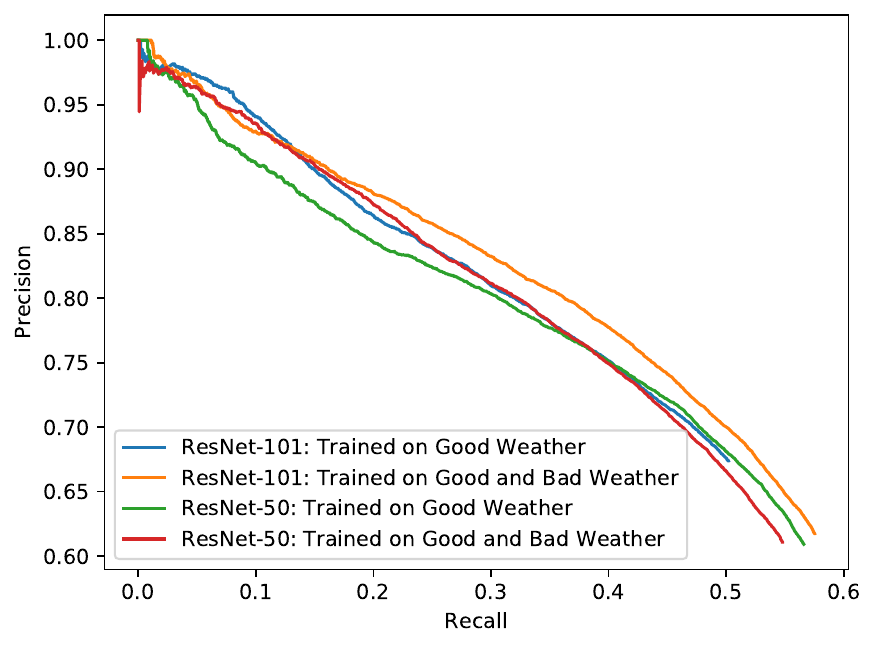}
\end{center}
    \caption{Precision recall curves.}\label{fig:prec_call_rot}
\end{figure}

\begin{figure}[t!]
    \begin{center}
    \subfloat[Train with good weather only.]{%
    \label{fig:heat_good}%
    \includegraphics[width=0.47\textwidth]{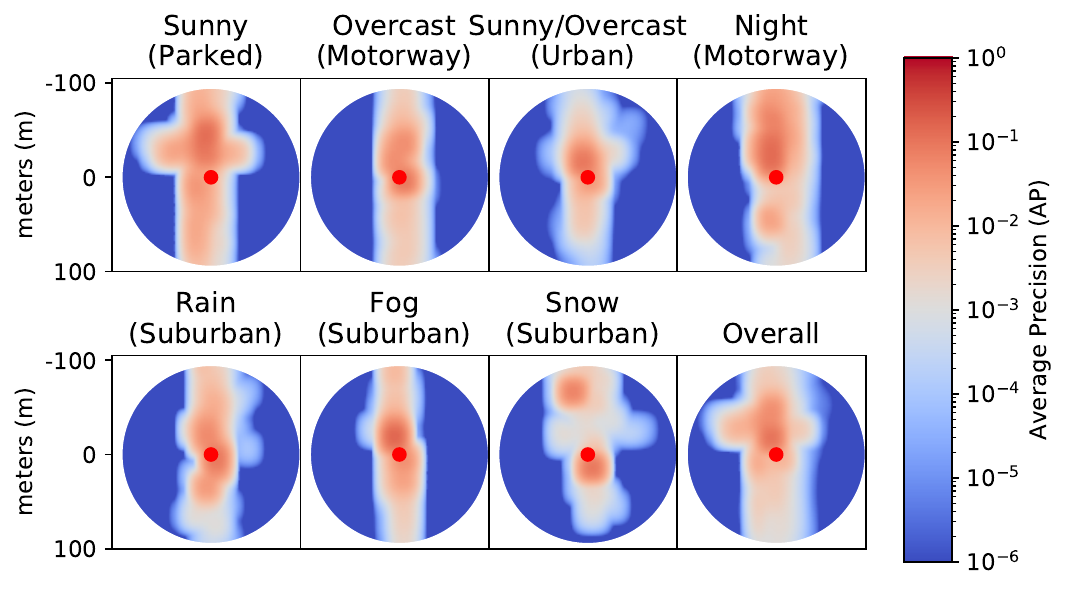}}%
    \qquad
    \subfloat[Train with both good and bad weather.]{%
    \label{fig:heat_gab}%
    \includegraphics[width=0.47\textwidth]{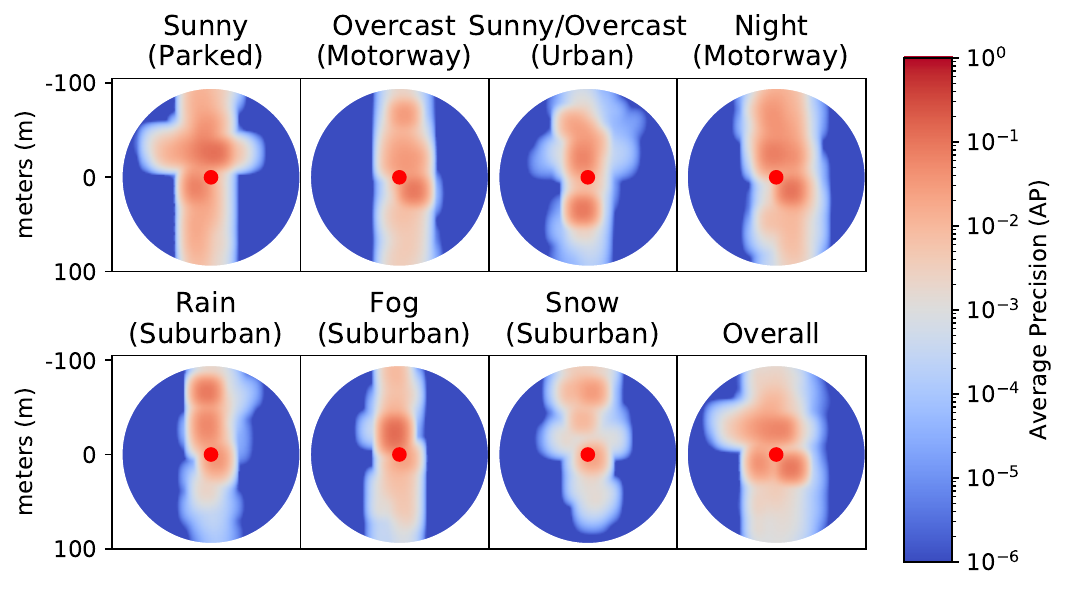}}%
    \end{center}
    \caption{Heatmap of Average Precision.}\label{fig:heatmap}
\end{figure}

\begin{figure}[t!]
    \begin{center}
    \includegraphics[width=0.75\textwidth]{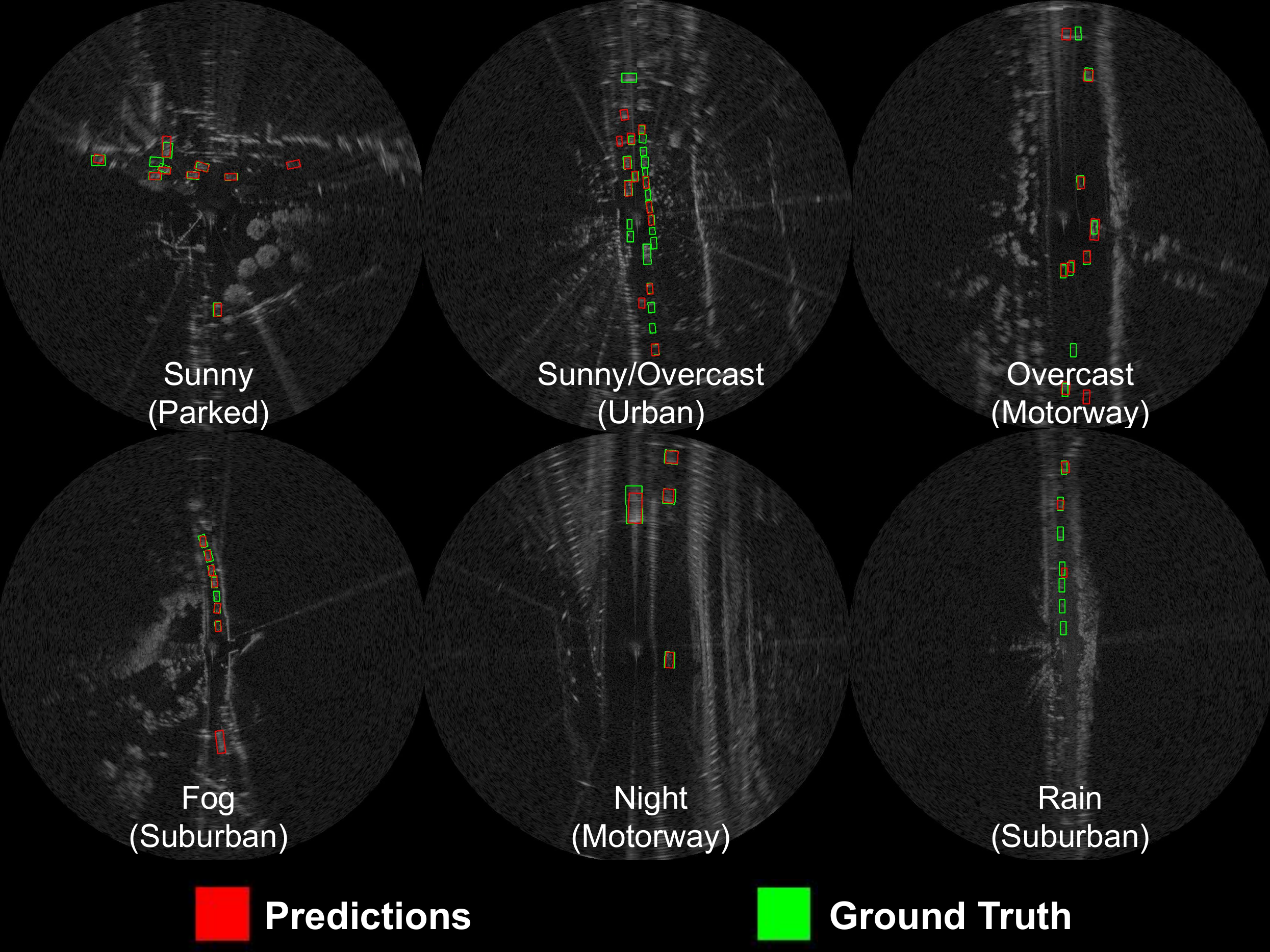}
    \end{center}
    \caption{Qualitative results of radar based vehicle detection.}
    \label{fig:qual_rot}
    
\end{figure}

To investigate the impact of weather conditions, the models were trained with the 2 different training datasets: data from only good and data from both good and bad weather. ResNet-50 and ResNet-101 were chosen as backbone models \cite{he2016deep}. The trained models were tested on a test set collected from all weather conditions and driving scenarios.  The metric used for evaluation was Average Precision with Intersection over Union (IoU) equal to 0.5, which is the same as the PASCAL VOC \cite{everingham2010pascal} and DOTA \cite{xia2018dota} evaluation metrics. Table \ref{tab:ap_rot} shows the Average Precision (AP) results and Figure \ref{fig:prec_call_rot} shows the precision recall curve. It can be seen that the AP difference between training with good weather and good\&bad weathers is marginal, which suggests that the weather conditions cause no or only a subtle impact on radar based object detection. The heat maps of the AP with respect to the radar image coordinates are also given in Figure \ref{fig:heatmap}. Since the AP distribution of the model trained only on the good weather data is very similar to the model trained with both good and bad weather data, it further verifies that radar is promising for object detection in all weathers. Regarding the results in each scenario, it is mainly biased by the type of data, rather than the weather itself. The parked scenario is shown to be the easiest, achieving close to 80\% AP, potentially aided by the consistency in the radar return from the environmental surround. Results in snow and rain data performed more poorly. Examining Figure \ref{fig:lidar_fog}, the radar sensor used was affected by rain changing the background pixel values.
In fog, we achieved considerably better results using radar. Since it is a challenging scenario for optical sensors, radar is shown to be a good solution for dense fog perception. In nighttime, motorway scenarios, the results were close to the results in daytime. This was expected, since radar is an active sensor and is not affected by the lack of illumination. Figure \ref{fig:qual_rot} illustrates some qualitative results for radar based vehicle detection in various driving scenarios and weather conditions, using the Faster R-CNN ResNet-101 trained in good weather only.

\begin{table}
\captionof{table}{Average Precision results on test set.}\label{tab:ap_rot}
\begin{center}
\def\arraystretch{1}%
\resizebox{\columnwidth}{!}{\begin{tabular}{l|c|ccccccc}
{} & Overall & \makecell{Sunny \\(Parked)} & \makecell{Overcast\\(Motorway)} & \makecell{Sunny/Overcast\\(Urban)} & \makecell{Night \\ (Motorway)} &  \makecell{Rain \\ (Suburban)} &   \makecell{Fog \\ (Suburban)} &  \makecell{Snow \\ (Suburban)} \\
\hline
ResNet-50 Trained on Good and Bad Weather &   45.77 &  78.99 &    42.06 & 36.12 & 54.71 & 33.53 & 48.24 & 12.81 \\
ResNet-50 Trained on Good Weather        &   45.31 &  78.15 &    47.06 & 37.04 & 51.80 & 26.45 & 47.25 &  5.47 \\
ResNet-101 Trained on Good and Bad Weather &   46.55 &  79.72 &    44.23 & 35.45 & 64.29 & 31.96 & 51.22 &  8.14 \\
ResNet-101 Trained on Good Weather   &   45.84 &  78.88 &    41.91 & 30.36 & 40.49 & 29.18 & 48.30 & 11.16 \\
\end{tabular}}
\end{center}

\end{table}

\section{Conclusions}

We have presented the new, labelled, large scale RADIATE dataset for research into sensory data perception for autonomous and assisted driving. The sensor setup, calibration and labelling processes are described, and we show examples of data collected under several different weather and road scenarios. As an example of use, we show how it can be used for radar based vehicle detection in the wild. These results show that radar based object detection is less affected by the weather, especially in foggy scenarios, where recognition from LiDAR data fails at a very short range depending on the fog density. This initial, baseline experiment demonstrated promising results, especially in adverse conditions. 
In this preliminary study, we emphasise that motion is not used for recognition, in part as actors may well be stationary, and in part because no Doppler data is available, although frame to frame tracking is an obvious avenue for further research provided a long term motion memory model is employed. We hope RADIATE will facilitate research in the community, particularly for robust perception in adverse weather.

\section*{Acknowledgements} 
This work was supported by Jaguar Land Rover and the UK Engineering and Physical Research Council, grant reference EP/N012402/1 (TASCC: Pervasive low-TeraHz and Video Sensing for Car Autonomy and Driver Assistance (PATH CAD)). We would like to thank Georgios Kalokyris for helping with the data annotation. We also thank NVIDIA for the donation of the TITAN X GPU.

\pagebreak

\bibliographystyle{IEEEtran}
\bibliography{egbib}

\begin{thebibliography}{10}
\providecommand{\url}[1]{#1}
\csname url@samestyle\endcsname
\providecommand{\newblock}{\relax}
\providecommand{\bibinfo}[2]{#2}
\providecommand{\BIBentrySTDinterwordspacing}{\spaceskip=0pt\relax}
\providecommand{\BIBentryALTinterwordstretchfactor}{4}
\providecommand{\BIBentryALTinterwordspacing}{\spaceskip=\fontdimen2\font plus
\BIBentryALTinterwordstretchfactor\fontdimen3\font minus
  \fontdimen4\font\relax}
\providecommand{\BIBforeignlanguage}[2]{{%
\expandafter\ifx\csname l@#1\endcsname\relax
\typeout{** WARNING: IEEEtran.bst: No hyphenation pattern has been}%
\typeout{** loaded for the language `#1'. Using the pattern for}%
\typeout{** the default language instead.}%
\else
\language=\csname l@#1\endcsname
\fi
#2}}
\providecommand{\BIBdecl}{\relax}
\BIBdecl

\bibitem{Geiger2013IJRR}
A.~Geiger, P.~Lenz, C.~Stiller, and R.~Urtasun, ``Vision meets robotics: The
  kitti dataset,'' \emph{International Journal of Robotics Research (IJRR)},
  2013.

\bibitem{nuscenes2019}
H.~Caesar, V.~Bankiti, A.~H. Lang, S.~Vora, V.~E. Liong, Q.~Xu, A.~Krishnan,
  Y.~Pan, G.~Baldan, and O.~Beijbom, ``nuscenes: A multimodal dataset for
  autonomous driving,'' \emph{arXiv preprint arXiv:1903.11027}, 2019.

\bibitem{sun2019scalability}
P.~Sun, H.~Kretzschmar, X.~Dotiwalla, A.~Chouard, V.~Patnaik, P.~Tsui, J.~Guo,
  Y.~Zhou, Y.~Chai, B.~Caine \emph{et~al.}, ``Scalability in perception for
  autonomous driving: An open dataset benchmark,'' \emph{arXiv preprint
  arXiv:1912.04838}, 2019.

\bibitem{pfennigbauer2014online}
M.~Pfennigbauer, C.~Wolf, J.~Weinkopf, and A.~Ullrich, ``Online waveform
  processing for demanding target situations,'' in \emph{Laser Radar Technology
  and Applications XIX; and Atmospheric Propagation XI}, vol. 9080.\hskip 1em
  plus 0.5em minus 0.4em\relax International Society for Optics and Photonics,
  2014, p. 90800J.

\bibitem{SDV18}
\BIBentryALTinterwordspacing
C.~Sakaridis, D.~Dai, and L.~Van~Gool, ``Semantic foggy scene understanding
  with synthetic data,'' \emph{International Journal of Computer Vision}, vol.
  126, no.~9, pp. 973--992, Sep 2018. [Online]. Available:
  \url{https://doi.org/10.1007/s11263-018-1072-8}
\BIBentrySTDinterwordspacing

\bibitem{bijelic2018benchmark}
M.~Bijelic, T.~Gruber, and W.~Ritter, ``A benchmark for lidar sensors in fog:
  Is detection breaking down?'' in \emph{2018 IEEE Intelligent Vehicles
  Symposium (IV)}.\hskip 1em plus 0.5em minus 0.4em\relax IEEE, 2018, pp.
  760--767.

\bibitem{kutila2018automotive}
M.~Kutila, P.~Pyyk{\"o}nen, H.~Holzh{\"u}ter, M.~Colomb, and P.~Duthon,
  ``Automotive lidar performance verification in fog and rain,'' in \emph{2018
  21st International Conference on Intelligent Transportation Systems
  (ITSC)}.\hskip 1em plus 0.5em minus 0.4em\relax IEEE, 2018, pp. 1695--1701.

\bibitem{WalHalBul2020}
\BIBentryALTinterwordspacing
A.~M. Wallace, A.~Halimi, and G.~S. Buller, ``Full waveform lidar for adverse
  weather conditions,'' \emph{IEEE Transactions on Vehicular Technology}, vol.
  Early Access, 2020. [Online]. Available:
  \url{https://ieeexplore.ieee.org/document/9076331}
\BIBentrySTDinterwordspacing

\bibitem{daniel2017low}
L.~Daniel, D.~Phippen, E.~Hoare, A.~Stove, M.~Cherniakov, and M.~Gashinova,
  ``Low-thz radar, lidar and optical imaging through artificially generated
  fog,'' in \emph{International Conference on Radar Systems (Radar
  2017)}.\hskip 1em plus 0.5em minus 0.4em\relax IET, 2017, pp. 1--4.

\bibitem{norouzian2019experimental}
F.~Norouzian, E.~Marchetti, E.~Hoare, M.~Gashinova, C.~Constantinou,
  P.~Gardner, and M.~Cherniakov, ``Experimental study on low-thz automotive
  radar signal attenuation during snowfall,'' \emph{IET Radar, Sonar \&
  Navigation}, vol.~13, no.~9, pp. 1421--1427, 2019.

\bibitem{norouzian2019rain}
F.~Norouzian, E.~Marchetti, M.~Gashinova, E.~Hoare, C.~Constantinou,
  P.~Gardner, and M.~Cherniakov, ``Rain attenuation at millimeter wave and
  low-thz frequencies,'' \emph{IEEE Transactions on Antennas and Propagation},
  vol.~68, no.~1, pp. 421--431, 2019.

\bibitem{Chang_2019_CVPR}
M.-F. Chang, J.~Lambert, P.~Sangkloy, J.~Singh, S.~Bak, A.~Hartnett, D.~Wang,
  P.~Carr, S.~Lucey, D.~Ramanan, and J.~Hays, ``Argoverse: 3d tracking and
  forecasting with rich maps,'' in \emph{The IEEE Conference on Computer Vision
  and Pattern Recognition (CVPR)}, June 2019.

\bibitem{RadarRobotCarDatasetICRA2020}
\BIBentryALTinterwordspacing
D.~Barnes, M.~Gadd, P.~Murcutt, P.~Newman, and I.~Posner, ``The oxford radar
  robotcar dataset: A radar extension to the oxford robotcar dataset,'' in
  \emph{Proceedings of the IEEE International Conference on Robotics and
  Automation (ICRA)}, Paris, 2020. [Online]. Available:
  \url{https://arxiv.org/abs/1909.01300}
\BIBentrySTDinterwordspacing

\bibitem{gskim-2020-mulran}
G.~Kim, Y.~S. Park, Y.~Cho, J.~Jeong, and A.~Kim, ``Mulran: Multimodal range
  dataset for urban place recognition,'' in \emph{Proceedings of the IEEE
  International Conference on Robotics and Automation (ICRA)}, Paris, May 2020,
  accepted. To appear.

\bibitem{meyer2019automotive}
M.~Meyer and G.~Kuschk, ``Automotive radar dataset for deep learning based 3d
  object detection,'' in \emph{2019 16th European Radar Conference
  (EuRAD)}.\hskip 1em plus 0.5em minus 0.4em\relax IEEE, 2019, pp. 129--132.

\bibitem{tiradar}
\BIBentryALTinterwordspacing
\emph{Short Range Radar Reference Design Using AWR1642}, April 2017. [Online].
  Available: \url{http://www.ti.com/lit/ug/tidud36b/tidud36b.pdf}
\BIBentrySTDinterwordspacing

\bibitem{SheWalWan2020}
M.~Sheeny, A.~Wallace, and S.~Wang, ``300 ghz radar object recognition based on
  deep neural networks and transfer learning,'' \emph{IET Proceedings on Radar,
  Sonar and Navigation}, 2020.

\bibitem{sheeny2020radio}
------, ``Radio: Parameterized generative radar data augmentation for small
  datasets,'' \emph{Applied Sciences}, vol.~10, no.~11, p. 3861, 2020.

\bibitem{Major_2019_ICCV}
B.~Major, D.~Fontijne, A.~Ansari, R.~Teja~Sukhavasi, R.~Gowaikar, M.~Hamilton,
  S.~Lee, S.~Grzechnik, and S.~Subramanian, ``Vehicle detection with automotive
  radar using deep learning on range-azimuth-doppler tensors,'' in \emph{The
  IEEE International Conference on Computer Vision (ICCV) Workshops}, Oct 2019,
  pp. 0--0.

\bibitem{Sless_2019_ICCV}
L.~Sless, B.~El~Shlomo, G.~Cohen, and S.~Oron, ``Road scene understanding by
  occupancy grid learning from sparse radar clusters using semantic
  segmentation,'' in \emph{The IEEE International Conference on Computer Vision
  (ICCV) Workshops}, Oct 2019.

\bibitem{wang2020rodnet}
Y.~Wang, Z.~Jiang, X.~Gao, J.-N. Hwang, G.~Xing, and H.~Liu, ``Rodnet: Object
  detection under severe conditions using vision-radio cross-modal
  supervision,'' \emph{arXiv preprint arXiv:2003.01816}, 2020.

\bibitem{ros}
M.~Quigley, K.~Conley, B.~P. Gerkey, J.~Faust, T.~Foote, J.~Leibs, R.~Wheeler,
  and A.~Y. Ng, ``Ros: an open-source robot operating system,'' in \emph{ICRA
  Workshop on Open Source Software}, 2009.

\bibitem{velodyne}
Velodyne, ``Velodyne hdl-32e,''
  \url{https://velodynelidar.com/products/hdl-32e/}, 2020.

\bibitem{navtech_radar}
Navtech, ``Navtech radar technical specifications,''
  \url{https://navtechradar.com/clearway-technical-specifications/}, [Online;
  accessed April 03, 2020].

\bibitem{matlabtool}
``Matlab stereo camera calibration,'' \url{
  https://uk.mathworks.com/help/vision/ref/stereocameracalibrator-app.html},
  2020, [Online; accessed April 03, 2020].

\bibitem{cvat}
``Cvat: Powerful and efficient computer vision annotation tool,''
  \url{https://github.com/opencv/cvat}.

\bibitem{vatic}
\BIBentryALTinterwordspacing
C.~Vondrick, D.~Patterson, and D.~Ramanan, ``Efficiently scaling up
  crowdsourced video annotation,'' \emph{International Journal of Computer
  Vision}, pp. 1--21, 2013. [Online]. Available:
  \url{http://dx.doi.org/10.1007/s11263-012-0564-1}
\BIBentrySTDinterwordspacing

\bibitem{labelimg}
``Labelimg,'' \url{https://github.com/tzutalin/labelImg}, [Online; accessed
  April 03, 2020].

\bibitem{allen2004object}
J.~G. Allen, R.~Y. Xu, J.~S. Jin \emph{et~al.}, ``Object tracking using
  camshift algorithm and multiple quantized feature spaces,'' in \emph{ACM
  International Conference Proceeding Series}, vol. 100, 2004, pp. 3--7.

\bibitem{ren2015faster}
S.~Ren, K.~He, R.~Girshick, and J.~Sun, ``Faster r-cnn: Towards real-time
  object detection with region proposal networks,'' in \emph{Advances in neural
  information processing systems}, 2015, pp. 91--99.

\bibitem{he2016deep}
K.~He, X.~Zhang, S.~Ren, and J.~Sun, ``Deep residual learning for image
  recognition,'' in \emph{Proceedings of the IEEE Conference on Computer Vision
  and Pattern Recognition}, 2016, pp. 770--778.

\bibitem{everingham2010pascal}
\BIBentryALTinterwordspacing
M.~Everingham, L.~Van~Gool, C.~K. Williams, J.~Winn, and A.~Zisserman, ``The
  pascal visual object classes (voc) challenge,'' \emph{International journal
  of computer vision}, vol.~88, no.~2, pp. 303--338, Jun 2010. [Online].
  Available: \url{https://doi.org/10.1007/s11263-009-0275-4}
\BIBentrySTDinterwordspacing

\bibitem{xia2018dota}
G.-S. Xia, X.~Bai, J.~Ding, Z.~Zhu, S.~Belongie, J.~Luo, M.~Datcu, M.~Pelillo,
  and L.~Zhang, ``Dota: A large-scale dataset for object detection in aerial
  images,'' in \emph{Proceedings of the IEEE Conference on Computer Vision and
  Pattern Recognition}, 2018, pp. 3974--3983.

\end{thebibliography}

\end{document}